\documentclass[runningheads]{llncs}
\usepackage[T1]{fontenc}
\pdfoutput=1

\usepackage{graphicx}

\usepackage{amsfonts}
\usepackage[many]{tcolorbox}    
\usepackage{enumitem}
\usepackage[sorting=nty]{biblatex}
\definecolor{sub}{HTML}{cde4ff}     % setting sub color to be used

\usepackage[hyperindex,breaklinks]{hyperref}
\hypersetup{colorlinks,
            allcolors=black,
            breaklinks=true}

\DeclareSymbolFont{largesymbolsA}{U}{txexa}{m}{n}
\DeclareMathSymbol{\varprod}{\mathop}{largesymbolsA}{16}

\tcbset{
    sharp corners,
    colback = white,
    before skip = 0.2cm,
}                           

\newtcolorbox{boxC}{
    rounded corners,
    arc = 6pt,
    colback = sub, 
    colframe = sub, 
    boxrule = 0pt, 
}

\begin{document}
\title{
Quantitative Assessment of Intersectional Empathetic Bias and Understanding}
\titlerunning{Quantitative Assessment of Empathetic Understanding}
\author{Vojtěch Formánek\inst{1, 2}
        \and Ondřej Sotolář\inst{1}}
\authorrunning{V. Formánek\and O. Sotolář}
\institute{Masaryk University, Faculty of Informatics \and Masaryk University, Department of Psychology, Faculty of Arts 
\email{xforman@mail.muni.cz}}
\maketitle             

\begin{abstract}
A growing amount of critique concerns the current operationalizations of empathy based on loose definitions of the construct. Such definitions negatively affect dataset quality, model robustness, and evaluation reliability. We propose an empathy evaluation framework that operationalizes empathy close to its psychological origins. The framework measures the variance in responses of LLMs to prompts using existing metrics for empathy and emotional valence. We introduce the variance by varying social biases in the prompts, which affect context understanding and thus impact empathetic understanding. Our method maintains high control over the prompt generation, ensuring the theoretical validity of the constructs in the prompt dataset. Also, it makes high-quality translation, especially into languages with little to no way of evaluating empathy or bias, such as the Slavonic family, more manageable. Using chosen LLMs and various prompt types, we demonstrate the empathy evaluation with the framework, including multiple-choice answers and free generation. The measured variance in our initial evaluation sample is small, and we were unable to find the expected differences between the empathetic understanding given the differences in context for distinct social groups. However, the models showed significant alterations in their reasoning chains that were needed to capture the relatively subtle changes in the prompts. This provides the basis for future research into the construction of the evaluation sample and statistical methods for measuring the results. 

\keywords{Intersectional Bias  \and Empathy \and LLM Evaluation}
\end{abstract}
\section{Introduction\label{sec:introduction}}
While there has been a vast amount of literature on empathy, it has come under increased scrutiny due to the unclear way of operationalizing empathy~\cite{lee2024comparative, lahnala2022critical}. Loosely defining empathy as \textit{the ability to understand another person's feelings and respond appropriately}~\cite{lahnala2022critical} was shown to cause problems across different tasks, such as dataset creation~\cite{debnath-2023-critical}, training~\cite{sotolar2024empoemotiongroundingempathetic}, and evaluation~\cite{lahnala2022critical}. This ambiguity led to a narrow focus on emotion recognition and prediction. We argue, alongside previous work, that this effort misunderstands empathy's psychological origins. To improve upon the current operationalizations of empathy, we propose a framework with (i) disambiguation of empathy, (ii) measurement operationalization intended specifically for computational models and (iii) an evaluatory procedure.

Some of the problems stem from the disagreement on the definition of empathy within psychology itself~\cite{coll2017we}. It was originally used to describe human ability to understand others~\cite{lipps1909leitfaden, jahoda2005theodor}. Current research agrees that it has two components to achieve this: \textit{affective empathy}, sometimes also named emotional empathy, and \textit{cognitive empathy}~\cite{spaulding2017cognitive}. Affective empathy refers to the capacity to feel emotions for others as a result of our belief, perception or imagination of their situation~\cite{maibom2017affective}. Cognitive empathy involves theorizing about and simulation of others mental states~\cite{spaulding2017cognitive}, that is to: (i) retrodictively simulate a mental state, to explain observed behavior, (ii) take that mental state and run it through our cognitive mechanisms, and (iii) attribute the conclusion to the target for explanation and prediction.

%% relation of empathy to experience
Since cognitive empathy is dependent on one's cognitive mechanisms, it is also dependent on experience. Because of that, a person can have different levels of understanding based on the similarity of their experience and the state they are observing and feel different levels of empathy toward different social groups~\cite{Zhao2023-so}. This effect is carried over to LLMs~\cite{santurkar2023whose, Raj2024-ik, stade2023artificial, chiu2024computational} which are reinforced on human preference data. Observing if models exhibit this type of bias toward some of the groups thus can be leveraged to indirectly study empathy. 

We propose an empathy evaluation framework for conversational agents, such as LLMs, which focuses on empathetic understanding. The framework uses masked templates to generate an evaluation dataset of prompts designed for the agents to respond to. 
The templates include masked sections into which different information is inserted. The information is biased towards different social groups; the selection of the type of information, the values, and social groups are inspired by current research such as reviewed in Gallegos et al.~\cite{gallegos2024bias}.

Filling the masked section with varying values results in the evaluation sample. This enables measuring the variance in responses across a single template, which assumes the invariance in empathetic understanding, affect, and responding both inside and at the intersection of different social groups. This invariance also means that bias within the framework, observed in the scores of a given metric, manifests as a deviation from the central tendency of the scores across a given social group. 

This might or might not be preferable; thus, we give fine-grained tools for individually interpreting the scores within a given intersectional group. Taking into account Blodgett et al.'s~\cite{blodgett2020language} criticism of the study of biases, within the framework, we only study the tendencies in model outputs and make no claims about the potential harmfulness or possible impacts of the biases. We make all the data and code publicly available\footnote{\url{https://github.com/xforman/JaEm_st}}.

\section{Related Work\label{sec:related_work}}
Several metrics have been proposed for evaluating bias in generated text~\cite{gallegos2024bias}. The metrics can be based on a difference in the distribution of tokens in the generated outputs given distinct groups \cite{rajpurkar-etal-2016-squad} or on lexicons \cite{dhamala2021bold}. Alternatively, classifiers are utilized, typically to detect relevant phenomena such as toxicity~\cite{sheng2019woman, sicilia2023learning}. The datasets used for evaluating bias deal with various social groups and issues (see Table 4 in~\cite{gallegos2024bias} for an overview) and are sometimes created from existing datasets. Sample construction involves replacing the relevant social group identifiers $G_1, ...,G_m \in \mathbb{G}$ (gender, race, etc.) with a mask, thus creating masked samples. When evaluating a mask, masking a specific social group $G_i$, it is substituted with a protected attribute $a_i\in G_i$ from that social group. The shift in the responses is measured~\cite{morales2024dsl, wan2023biasasker}, assuming that the output should be invariant considering distinct social groups. This technique is sometimes called bias mitigation via contact hypothesis~\cite{Raj2024-ik}, a term borrowed from psychology referring to direct contact with other social groups~\cite{paluck2019contact}.  

Datasets used to train empathetic agents are typically single or multi-turn, with emotional labels~\cite{buecheletal2018modeling, rashkin-etal-2019-towards}. However, other datasets, by their nature, contain empathy as well, such as transcripts of everyday conversations~\cite{Li2017-za}, simulation of other's personas~\cite{Zhang2018-uq} or transcripts of therapies~\cite{malhotra2022speaker, perez2019makes}, labeled with conversational behaviors~\cite{chiu2024computational}. Retroactively categorizing existing empathy metrics into the two dimensions is difficult, especially since they likely overlap.

Since \textit{cognitive} empathy involves understanding, the accuracy of emotion prediction can be considered a case of it, but a broader understanding has also been measured. Zhu et al.~\cite{zhu2024reading} collected user comments about products and their do-, motor- and be-goals~\cite{hassenzahl2010experience}, then instructed human or LLM designers to predict those goals and measured the token similarity between them. 

The problems with measuring \textit{affective} empathy are caused mainly by its dependence on an inner state. Lee et al.~\cite{lee2024comparative} propose a set of feature-based metrics for evaluating model responses, which include mechanisms outside empathy, but we consider them relevant because they measure the empathetic qualities indirectly. Concretely, the set includes specificity, based on a normalized variant of inverse document frequency (NIDF)~\cite{see2019makes} and measures the similarity in the vocabularies of the model and user. They also introduce valence, arousal, and dominance (VAD) based on the NRC Emotion Intensity Lexicon~\cite{mohammad2017word}. With the focus on the similarity of the input and output texts, closer results are preferred. All of the \textit{affective} metrics assume that empathy in this context manifests in the similarity. The described metrics also fall into the category of empathetic responding, which is the focus of many currently existing measures. A different method is used in the \textsc{Epitome} empathy metric for dialogues~\cite{sharma-etal-2020-computational}, which is based on a fine-tuned RoBERTa model that predicts three dimensions on a scale (0-2; none, weak, strong): Empathetic Responses (ER), Explanations (EX) and Interpretations (IP). Chiu et al.~\cite{chiu2024computational} evaluate differences between human and LLM therapists and define several dimensions whose quality is in part dependent on the empathetic capacity of the therapist -- Reflections, Questions, and Solutions. 

All of the metrics depend on the true state of the evaluated sample (for example, the labeled emotion). This makes both dimensions of empathy dependent on this state, meaning that misunderstanding leads to an inaccurate affect. Thus, we cannot separate \textit{affective} from \textit{cognitive} empathy. For this reason, Coll et al.~\cite{coll2017we} operationalizes the measurement of \textit{affective} as the similarity of that affective state to the one in the \textit{understood} state. 

\section{JaEm-ST: Framework for the Quantitative Assessment of Empathetic Behavior of LLMs}
We propose an evaluation framework, JaEm-ST, where empathy has two dimensions, \textit{Cognitive} and \textit{Affective}, which follow the definitions introduced in Section~\ref{sec:introduction}. However, we operationalize the measurement into three dimensions. Cognitive empathy (CE), the degree to which the empathizer understood the observed state correctly. Affective empathy (AE), the degree to which the empathizer's state matches that of the understood state (inspired by Coll et al.~\cite{coll2017we}). Lastly, we define \textit{Empathic Response Appropriateness} (ERA) to the understood state, which is the result of the empathic process (and several others) but not its dimension. In our context, we define ``state'' as a person's momentary mental and physical circumstances. 

\subsection{Theoretical Basis of the Framework}
Dependence of \textit{cognitive} empathy on experience means that different empathizers might come to different conclusions about an observed state, even when self-reporting~\cite{grainger2023measuring}. This impacts the evaluation on two sides: (i) LLM empathizers might interpret the situation differently, which does not mean that it is false, and (ii) the ``true state'' constructed by the creator might not even be reflective of their empathetic understanding. Thus, it is difficult, if not impossible, to determine whether a given interpretation is genuinely false, but it is nonetheless representative of a human interpretation grounded in experience. For this reason, AE and ERA depend on the model's understanding of the state, so it is possible to evaluate an output even when it does not interpret the context in the same way as the creator of that template. But if we concede that the interpretation of states is subjective, then we need another standard for the evaluation. 

In our case we assume that empathetic understanding is invariant across similar situations; the implications of this assumption are discussed later. Which is why JaEm-ST focuses on finding systemic differences in model output when responding to similar situations and uses the similarity between the ``true state'' and the one predicted by the model as a guiding principle. Given that experience can lead to biases, we create a single evaluation example by inserting protected attributes $a_i = (a_{i1}, ..., a_{ih}) \in G_1, ..., G_h$ (such as specific sexuality, education etc.)  associated with their respective social groups $G_k\in\mathbb{G}$ into a masked template $t\in \mathbb{T}$ (see Fig~\ref{fig:answer_example}), by a procedure $fill(t, a)$. Thus two examples created from the same template differ only in the specific protected attributes that were inserted into them.

\subsection{Evaluation Sample\label{sec:eval_sample}}
We create the evaluation sample $D$ dynamically from a small seed dataset of predefined templates $\mathbb{T}$. The templates simulate conversations between a human speaker and an empathizer, which is assumed to be a conversational agent and is to be evaluated. It consists of four parts: instruction, context, conversation, and answers. The templates are then filled with different combinations of protected attributes $A \subseteq \varprod^{h}_{k=1} G_k$, thus $D = \{fill(t, a)~|~\forall t_{\in \mathbb{T}}, a_{\in A}\}$. For us $h = 7$ and $A$ is a set of commonly examined social groups: race, education, religion, age~\cite{morales2024dsl} and socio-economic status~\cite{wan2023biasasker}, sexuality and pronouns. Typically, the cartesian product of all possibilities, i.e. $A=\varprod^{h}_{k=1} G_k$, would be too large, thus we only fully sample the dimensions of interest and randomly sample the rest.

The templates are based on \textit{causal tuples}, which provide the reasoning for what caused a person's current observed state. It is partly set up by the context, which defines the social groups the observed person is a member of. The conversation within the templates implicitly or explicitly describes the situation grounded on the causal tuple. The conversation can also include the same masks as the context. 

Given a filled template, the evaluated model is then prompted to continue the conversation. If evaluating AE and ERA, it simply continues generating the response to the observed person's last utterance. When evaluating CE, it is given a choice between five answers and is prompted to pick the one with the deepest understanding of how the person reached the observed state in the last utterance.

Even though we assume that empathetic understanding is invariant in the templates, there may be cases when this assumption does not hold. If certain types of responses are preferable for a given group, and this type of response is measured by a metric, it is possible to set thresholds for this metric and group separately, independently of the overall scores. This is possible because we have stratified information about the social groups which the observed person in the evaluated context is a part of. 

% ---------------------Example----------------------------
\begin{figure}
\begin{boxC}
\textbf{*Causal tuple} (not part of the model input in any case)\textbf{:} \\
(adores her parrot, happiness from a gift for her parrot, less critical towards disliked food, complimented the cook)\\

\textbf{*Prompt:}\\
When given a choice of answers A-E choose the one that shows the deepest understanding of Jane's situation, not necessarily the most empathetic. Carl has a full understanding of the context below:\\

\textbf{Context: }\\
Generally positive and optimistic person. [PRONOUN] is a [SEXUALITY] and a firm [RELIGION]. [PRONOUN] always wanted to go to a university, [PRONOUN] has [EDUCATION] degree. Being a proud cooking connoisseur and a stern critic, [PRONOUN] rarely compliments other's food, but today [PRONOUN] complimented pasta ... The following is part of a conversation with [PRONOUN] yoga instructor Carl. \\
\\
\textbf{Conversation:}\\
Carl: Thank you, I also liked the classes today, seems like a happy day for everyone. I wanted to make it a little harder for you. I noticed some exhausted faces, but not any annoyed ones. How were the exercises in the middle?\\
Jane: I didn't see any annoyed faces either ...\\
...\\
Jane: ... something weird happened to me today, we went for a typical lunch with my coworkers, but I think I lost my integrity, and complimented a food I did not actually like!\\ 
Carl: \\~\\
\textbf{*Answers:}\\
A: I think gift for Poppy made you think of her and made you so happy, that you complimented the chef even though you didn't actually like the food.\\
B: You? Did you actually complimented a food? you're always so strict and stern about critiquing other people's food. Maybe the day, you know, the yoga, Lucy's traveling and Poppy made you actually enjoy the food. What was it?\\
C: That's surprising, maybe it's because you were feeling so good ...  Anyhow, did Lucy tell you how her backpacking was in Texas?\\
D: Jane, I think it's understandable to feel like this. But your integrity isn't lost on an occasion like this. ... How do you feel about the situation now?\\
E: Don't be silly, it does not make you less of a critic. ...
\end{boxC}
    \centering
    \caption{An example of the framework's templates. [TAG] indicates masks to be replaced with a chosen social group's attributes. Answers and the prompt are only given to the model if cognitive empathy is being evaluated. Otherwise, it generates Carl's last response. Option (A) shows the deepest understanding since it is closest to the causal tuple -- three dots in the text mark parts excluded for brevity.}
    \label{fig:answer_example}
\end{figure}

%-------------------------------------------------------------

\subsection{Experimental setup}
For the evaluation, we constructed two causal tuples and templates , as shown in Figure~\ref{fig:answer_example}. To produce $D$, we focus only on two of the dimensions ($I$): \textbf{sexuality} and \textbf{pronoun}, we take all $a\in I$ four times and randomly sample from the rest of the dimensions. We get 486 samples for generation or multi-choice, so each model is evaluated with 972 samples, that is 1944 inputs in total.  

Because we aim to maintain an understandable, controllable, high quality dataset with high ecological validity, we keep the number of templates relatively low. This also means that it is relatively simple to translate the templates to other languages. This could be especially beneficial for languages that currently have little to no way of evaluating bias or empathy, such as the Slavonic family. 

The examples $d \in D$ are given as input to a model in a single-turn fashion, which is most typical for evaluating empathy. The examples are used as is, containing the following components in order: (i) $prompt_{CE}, context, conversation, \\answers$ for CE multiple choice, (ii) $context, conversation$ for AE and ERA. We pass each of the variants to the model separately. The resulting CE score is computed as the accuracy of selecting the choices that manifest the most understanding of the speaker's state across all samples. The AE score comprises Valence, Arousal, and Dominance scores. While we consider them unreliable measures of affective empathy because they measure it indirectly, they can provide valuable information about the type of affect the model uses in its response. For measuring ERA, we use the \textsc{Epitome}'s dimensions IP, EX, and ER.

We evaluate \textit{Llama-3.1-8B}~\cite{dubey2024llama} and \textit{Zephyr-gemma-v.1}~\cite{tunstall2023zephyr, zephyr_7b_gemma}. A 80 GB Nvidia A100 graphics card was used to process the evaluation.

\section{Results}
Table~\ref{tab:results} shows significant differences in all three evaluated dimensions (CE, AE, ERA) measured across the models and templates. The metrics respective to the dimensions are: accuracy of multiple-choice answers (CE), VAD (AE) and \textsc{Epitome} scores (ERA). For the different intersectional groups, the measured variance between scores is much smaller. 
\begin{table*}[ht]
  \centering
  \small
  \begin{tabular}{l|r|c|c|ccc|crl}
    \hline
      & & &\textit{Cognitive} & \multicolumn{3}{c|}{\textit{Affective}} & \multicolumn{3}{c}{\textit{Response}}\\
      & & &\textit{Empathy} & \multicolumn{3}{c|}{\textit{Empathy}} & \multicolumn{3}{c}{\textit{Appropriateness}}\\
      \textbf{Model}& \textbf{Template} & \textbf{Count} & \textbf{MC$\uparrow$} & \textbf{V$\downarrow$} & \textbf{A$\downarrow$} & \textbf{D$\downarrow$} & \textbf{IP$\uparrow$} & \textbf{EX$\uparrow$} & \textbf{ER$\uparrow$}\\
     \hline
    \textit{Llama-3.1}& 0 & 243 & $0.29$~ & $0.16$~ & $0.11$~ & $0.13$~ & $0.12$~ & $1.56$~ & $0.54$~ \\
                      & 1 & 243 & $0.35$~ & $0.15$~ & $0.11$~ & $0.13$~ & $0.19$~ & $0.09$~ & $0.91$~ \\                   
    \hline
    \textit{Zephyr-gemma} & 0 & 243 & $0.01$~ & $0.22$~ & $0.14$~ & $0.18$~ & $0.08$~ & $0.19$~ & $0.79$ \\
                          & 1 & 243 & $0.10$~ & $0.18$~ & $0.13$~ & $0.16$~ & $0.07$~ & $0.01$~ & $1.60$ \\
    \hline
  \end{tabular}
  \caption{Evaluation of the generated responses on the framework's three dimensions of empathy, the interpretation of the metrics is explained in Section~\ref{sec:related_work}. The metric for Cognitive empathy is the accuracy of selecting choices with the most understanding among the multi-choice answers (example in Fig.~\ref{fig:answer_example}). \textit{Llama-3.1} has higher accuracy in this task. \textit{Llama-3.1} is also more stable across the VAD metrics, and also Empathetic Explorations (EX) on only one of the templates, which were low otherwise. \textit{Zephyr-gemma} had higher scores in Empathetic Responding (ER), this may be caused by the fact that it tends to role-play less.}
  \label{tab:results}
\end{table*}
Generally, \textit{zephyr-gemma} performed much worse in CE. The results show that it was easiest for both of the models to find the answer with the most understanding to the sample shown in Figure~\ref{fig:answer_example}. The differences in AE scores between the models are smaller and \textit{Llama-3.1} achieves lower scores and smaller differences between the two samples, ERA scores are similar. 

For pronouns, one of the evaluated dimensions, there are no obvious differences between the scores. We found outliers in the intersection of these groups, such as Lesbian/She, which has an Interpretations (IP) score significantly above the overall average ($\sigma = 3.36$), but were unable to find any noticeable differences in the generated output.

Both models followed the multi-choice prompt well. For the free generation, manual inspection of a small subset the outputs suggests that \textit{Llama-3.1} might follow role-playing better, \textit{zephyr-gemma} tends respond from the third perspective (10-15 \%), or set the situation up in a couple of sentences (10 \%), instead of directly responding. 

\section{Conclusion}
We provided a disambiguation of empathy for computational models to help future work define the construct closer to its psychological origins as opposed to the loose definitions that are currently widespread. As a main result, we proposed a new empathy evaluation framework for the responses generated by conversational agents that acknowledges the inherent subjectivity of empathetic understanding. The framework focuses on how empathetic understanding and responding is influenced by intersectional bias. It provides methods to generate evaluation samples from templates by inserting the intersectional contexts into them. The framework uses a new three-dimensional measurement operationalization of empathy to measure the construct. We demonstrated the usage of the framework on a small synthetic sample. In all three framework dimensions, we measured significant differences between the \textit{Llama-3.1-8B} and \textit{Zephyr-gemma-v.1} models. Lastly, we identified differences in empathetic understanding across the evaluated metrics in some intersectional groups. More importantly, we showed the framework's strength in providing the ability to stratify scores across a wide range of social contexts, giving a more fine-grained insight into model behavior and potential harms.

\section*{Limitations}
We view the modest number of templates as a limitation. Even though we can produce many examples by substituting into the masks, most of their structure stays the same. Further, the contexts and conversations do not reflect the true variety across multiple different groups; future work should thus focus on increasing the number and diversity of the sample creators. The template structure itself, especially the inclusion of context as an explanation of the speaker's background, does place limitations the naturalness and ecological validity of the framework. Lastly, while the number of empathy metrics in this work is limited, future works can use the outlined criteria to include other metrics. 

\section*{Acknowledgments}
This work was supported by the project, Research of Excellence on Digital Technologies and Wellbeing \url{CZ.02.01.01/00/22_008/0004583} which is co-financed by the European Union.

\printbibliography
\end{document}